\pdfoutput=1

\documentclass[11pt]{article}

\usepackage[]{ACL2023}

\usepackage{times}
\usepackage{latexsym}

\usepackage[T1]{fontenc}

\usepackage{booktabs}
\usepackage{graphicx}
\usepackage{colortbl}
\usepackage{multirow}
\usepackage{bbding} 
\usepackage{makecell}
\usepackage{bm} 
\usepackage{color} 
\usepackage{xcolor} 
\usepackage{textcomp,booktabs} 
\usepackage{amsmath} 
\usepackage{amsfonts}
\usepackage{amsthm,amsmath,amssymb}
\usepackage{mathrsfs}
\usepackage{subfigure}
\usepackage{mathtools}
\usepackage[utf8]{inputenc}

\usepackage{microtype}

\usepackage{inconsolata}
\usepackage{CJKutf8}
\usepackage{makecell}

\newcommand{\secref}[1]{\S\ref{#1}}
\title{Disentangled Phonetic Representation for Chinese Spelling Correction}

\author{Zihong Liang\textsuperscript{\rm 1}, Xiaojun Quan\textsuperscript{\rm 1}\thanks{\;\;Corresponding authors}, Qifan Wang\textsuperscript{\rm 2} \\ \textsuperscript{\rm 1}School of Computer Science and Engineering, Sun Yat-sen University \textsuperscript{\rm 2}Meta AI \\ \texttt{liangzh63@mail2.sysu.edu.cn, quanxj3@mail.sysu.edu.cn, wqfcr@fb.com}
}

\begin{document}
\maketitle
\begin{abstract}
Chinese Spelling Correction (CSC) aims to detect and correct erroneous characters in Chinese texts. Although efforts have been made to introduce phonetic information (Hanyu Pinyin) in this task, they typically merge phonetic representations with character representations, which tends to weaken the representation effect of normal texts. In this work, we propose to disentangle the two types of features to allow for direct interaction between textual and phonetic information. To learn useful phonetic representations, we introduce a pinyin-to-character objective to ask the model to predict the correct characters based solely on phonetic information, where a separation mask is imposed to disable attention from phonetic input to text. To avoid overfitting the phonetics, we further design a self-distillation module to ensure that semantic information plays a major role in the prediction. Extensive experiments on three CSC benchmarks demonstrate the superiority of our method in using phonetic information\footnote{\url{https://github.com/liangzh63/DORM-CSC}}.

\end{abstract}

\section{Introduction}
\begin{CJK*}{UTF8}{gbsn}

Chinese Spelling Correction (CSC) is a task to detect and correct erroneous characters in Chinese sentences, which plays an indispensable role in many natural language processing (NLP) applications \citep{martins2004seq,gao2010large}. Previous research \citep{liu_visually} shows that the misuse of homophonic characters accounts for roughly 83\% of the spelling errors. 
\begin{table}[h]
    \centering
    \small
    \begin{tabular}{c|l}
    \toprule
    \multirow{2}{*}{Source} & 可是我忘了，我真\textcolor{red}{户秃(hu tu)}。  \\
                            & But I forgot, I am so household bald.  \\
    \\[-0.5em]
    \multirow{2}{*}{Target} & 可是我忘了，我真\textcolor{blue}{糊涂(hu tu)}。  \\
                            & But I forgot, I am so silly.  \\
    \hline
    \\[-0.9em]
    BERT        &  可是我忘了，我真\textcolor{red}{护突}。 \qquad \qquad \color{red}\XSolidBrush   \\
    PinyinBERT  &  可是我忘了，我真\textcolor{blue}{糊涂}。 \qquad \qquad \color{blue}\Checkmark  \\
    REALISE     &  可是我忘了，我真\textcolor{red}{户}\textcolor{blue}{涂}。 \qquad \qquad \color{red}\XSolidBrush  \\
    Our DORM  &  可是我忘了，我真\textcolor{blue}{糊涂}。\qquad \qquad \color{blue}\Checkmark   \\

    \hline
    \specialrule{0em}{2pt}{2pt}
    \hline
    \\[-0.9em]
    \multirow{2}{*}{Source} & 可是现在我什么事都不\textcolor{red}{济的(ji de)}。  \\
                            & But I can't do anything right now.  \\
    \\[-0.5em]
    \multirow{2}{*}{Target} & 可是现在我什么事都不\textcolor{blue}{记得(ji de)}。 \\
                            & But I don't remember anything now. \\
    \hline
    \\[-0.9em]
    BERT        &  可是现在我什么事都不\textcolor{blue}{记得}。 \qquad \color{blue}\Checkmark \\
    PinyinBERT  &  可是现在我什么事都不\textcolor{blue}{记}\textcolor{red}{的}。 \qquad \color{red}\XSolidBrush \\
    REALISE     &  可是现在我什么事都不\textcolor{blue}{记}\textcolor{red}{的}。 \qquad \color{red}\XSolidBrush\\
    Our DORM  &  可是现在我什么事都不\textcolor{blue}{记得}。 \qquad \color{blue}\Checkmark \\
    \bottomrule
    \end{tabular}
    \caption{Two examples of Chinese Spelling Correction and the predictions by different models. Misspelled characters are highlighted in red and the corresponding answers are in blue. The phonetic transcription of key characters is bracketed. PinyinBERT is a special BERT model which takes as input only phonetic features without characters. REALISE is a state-of-the-art model.}
    \label{tab:example}
    \vspace{-3mm}
\end{table}
We present two such cases in Table~\ref{tab:example}. In the first one, the erroneous characters of ``户秃'' are difficult to be corrected by only literal text because the input sample is too short and the two characters are entirely unrelated to the semantic meaning of this sample. However, their pronunciation easily helps us associate them with the correct answer ``糊涂'' which shares the same pronunciation as ``户秃''. The second case exhibits a similar phenomenon but is more complicated as the model must distinguish between ``记得'' and ``记的'' further. These two examples illustrate that misspelled characters could be recognized and corrected with the introduction of phonetic information. In Mandarin Chinese, Hanyu Pinyin (shortened to \emph{pinyin}) is the official romanization system for phonetic transcription. It uses three components of initials, finals, and tones to express the pronunciation and spelling of Chinese characters. As the pronunciation similarity of Chinese characters is primarily determined by their initial or final sounds rather than their tones, we focus solely on the initials and finals as the phonetic features of Chinese characters.

As pre-trained language models like BERT \citep{bert} have dominated various NLP tasks, researchers explore incorporating pinyin features into pre-trained language models for the CSC task. There are mainly two approaches. First, the pinyin of a Chinese character is encoded and fused into the character representation with a gate mechanism \citep{dcn, huangli, realise, mlmphonetics}. Second, a pronunciation prediction objective is introduced to model the relationship among phonologically similar characters \citep{plome, spellbert, scope}. 
Despite considerable performance gain, these methods suffer from two potential issues. First, pinyin information may be neglected or dominated by textual information during training because of the entanglement between pinyin and textual representations. As the first case shows in Table~\ref{tab:example}, a special BERT model taking only the pinyin sequence as input without Chinese characters can detect and correct the erroneous characters, while REALISE \citep{realise}, which encodes and fuses textual and pinyin information with a gate mechanism, ignores one of the errors. Second, the introduction of pinyin features may weaken the representation of normal texts. Take the second case in Table~\ref{tab:example} for example. While an ordinary BERT model can correct the misspelled character ``的'' in the input, REALISE fails to do that. This problem could be explained by the over-reliance of REALISE on or overfitting pinyin information.

Based on the above observations, we propose \textbf{D}isentangled ph\textbf{O}netic \textbf{R}epresentation \textbf{M}odel (DORM) for CSC. Our motivation is to decouple text and pinyin representations to allow for direct interaction between them to make better use of phonetic information. Specifically, we first construct a phonetics-aware input sequence by appending the pinyin sequence to the original textual input, where a common set of position embeddings is used to relate the two sub-sequences. In doing so, textual features are allowed to capture phonetic information as needed from the pinyin part during training and inference. Then, to learn useful pinyin representations, we introduce a pinyin-to-character prediction objective, where a separation mask is imposed to disallow attention from pinyin to text to ask the model to recover the correct characters only from pinyin information. The pinyin-to-character task is auxiliary during training and its prediction will be discarded at inference time. 

Intuitively, pinyin should serve to complement but not replace textual information in CSC for two reasons. First, there is a one-to-many relation between pinyin and Chinese characters, and it is more difficult to recover the correct characters solely from pinyin than from Chinese characters. Second, pinyin representations are not pre-trained as textual representations in existing language models. Therefore, the model should avoid overly relying on pinyin which may cause overfitting. Inspired by deep mutual learning \citep{mutual_learning} and self-distillation \citep{self_distillation}, we propose a self-distillation module to force the prediction of our model to be consistent with that when a raw-text input is supplied. To this end, KL-divergence is applied to the two sets of soft labels. 

Experiments are conducted on three SIGHAN benchmarks and the results show that our model achieves substantial performance improvement over state-of-the-art models. Further analysis demonstrates that phonetic information is better utilized in our model. The contributions of this work are summarized threefold. First, we disentangle text and pinyin representations to allow for direct interaction between them. Second, we introduce a pinyin-to-character task to enhance phonetic representation learning with a separation mask imposed to disable attention from pinyin to text. Third, a self-distillation module is proposed to prevent over-reliance on phonetic features. Through this work, we demonstrate the merit of our approach to modeling pinyin information separately from the text. 
\end{CJK*}

\section{Related Work}

\subsection{Chinese Spelling Correction}

Chinese Spelling Correction has drawn increasing interest from NLP researchers. The current methodology of this task has been dominated by neural network-based models, especially pre-trained language models, and can be divided into two lines. 

One line of work focuses on better semantic modeling of textual features \citep{faspell, gad, ecopo}. They treat CSC as a sequence labeling task and adopt pre-trained language models to acquire contextual representations. Soft-Masked BERT \citep{softmaskbert} employs a detection network to predict whether a character is erroneous and then generates soft-masked embedding for the correction network to correct the error. MDCSpell \citep{mdcspell} is a multi-task detector-corrector framework that fuses representations from the detection and correction networks. 

Another line of work is incorporating phonetic information into the task, motivated by the observation that the misuse of homophonic characters accounts for a large proportion of the errors \citep{liu_visually}. MLM-phonetics \citep{mlmphonetics} and PLOME \citep{plome} employ a word replacement strategy to replace randomly-selected characters with phonologically or visually similar ones in the pre-training stage. REALISE \citep{realise} and PHMOSpell \citep{huangli} utilize multiple encoders to model textual, phonetic, and visual features and employ a selective gate mechanism to fuse them. SCOPE \citep{scope} imposes an auxiliary pronunciation prediction task and devises an iterative inference strategy to improve performances. However, these methods generally merge textual and phonetic features without direct and deep interaction between them, which may lead to ineffective use of phonetic information. By contrast, our method decouples the two types of features to learn isolated phonetic representations and use them to assist textual information for CSC.

\subsection{Self-Distillation}

Knowledge distillation \citep{hinton2015distilling} is a technique that tries to distill a small student model from a large teacher model. As a special distillation strategy, deep mutual learning \citep{mutual_learning} allows several student models to collaboratively learn and teach each other during training. Particularly, it is referred to as self-distillation \citep{self_distillation} when the student models share the same parameters. Self-distillation has been applied in CSC and brings performance improvement. SDCL \citep{sdcl} encodes both original and corresponding correct sentences respectively, and adopts contrastive loss to learn better contextual representations. CRASpell \citep{liu-etal-2022-craspell} constructs a noisy sample for each input and applies KL-divergence for the two outputs to improve the performance on multi-typo sentences. Our method differs from CRASpell in two aspects. First, one of our student models takes as input a phonetics-aware sequence with disentangled textual and phonetic representations. Second, the purpose of our self-distillation design is to reduce overfitting phonetic information when training the model.

\section{Methodology}
\label{sec:methodology}
The motivation of our \textbf{D}isentangled ph\textbf{O}netic \textbf{R}epresentation \textbf{M}odel (DORM) for Chinese Spelling Correction (CSC) is to allow for direct and deep interaction between textual and phonetic features by decoupling Chinese character and pinyin representations. To enable effective pinyin representations, we introduce a pinyin-to-character objective that requires the model to restore the correct characters purely from pinyin information. Inspired by deep mutual learning \citep{mutual_learning} and self-distillation \citep{self_distillation}, we further introduce a self-distillation module to prevent the model from overfitting pinyin information. In the following, we first formulate the task (\secref{sec:problem_definition}) and then introduce DORM in detail (\secref{sec:model_detail}). Finally, we introduce how to pre-train the model for better textual and pinyin representations (\secref{subsec:pre-train}).

\subsection{Problem Definition}
\label{sec:problem_definition}
Given a Chinese sentence $X=\{x_1,x_2,..,x_n\}$ of $n$ characters that may include erroneous characters, we use $Y=\{y_1,y_2,..,y_n\}$ to represent the corresponding correct sentence. The objective of CSC is to detect and correct the erroneous characters by generating a prediction $\hat{Y}=\{\hat{y}_1,\hat{y}_2,..,\hat{y}_n\}$ for the input $X$, where $\hat{y}_i$ is the character predicted for $x_i$. Apparently, the CSC task can be formulated as a sequence labeling task in which all the Chinese characters constitute the label set.  

\begin{figure*}[t]
    \centering
    \includegraphics[width=\textwidth]{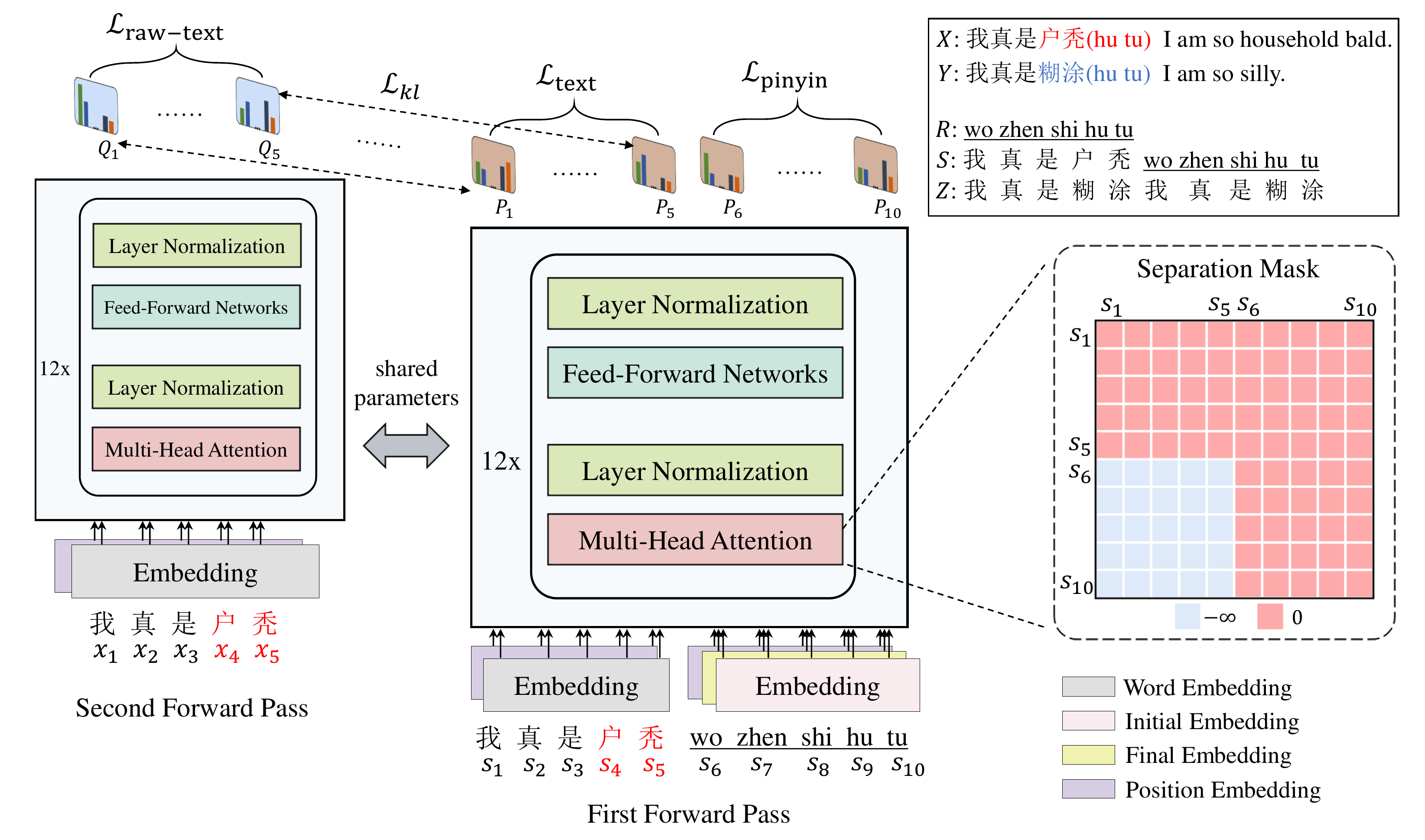}
    \caption{The architecture of the proposed DORM, which consists of a phonetics-aware input sequence $S$, an encoder with separation mask, a pinyin-to-character objective, and a self-distillation module. $X$ is the original input sentence, $R$ is the pinyin sequence of $X$, $Y$ is the corresponding correct sentence, and $Z$ is the prediction label based on $S$. Pinyin sequences are underlined to distinguish them from English sentences. Misspelled characters are shown in red and the corresponding correct characters are in blue. For self-distillation, the model conducts two forward passes with different inputs, and the output distributions are constrained by KL-divergence.}
    \label{fig:overview}
    \vspace{-3mm}
\end{figure*}

\subsection{Architecture}
\label{sec:model_detail}
As illustrated in Figure \ref{fig:overview}, our DORM consists of a phonetics-aware input sequence, a unified encoder with separation mask, a pinyin-to-character objective, and a self-distillation module. The phonetics-aware input is constructed by appending the pinyin sequence to the original textual input. The separation mask is imposed to disallow attention from pinyin to text to avoid information leaks. The pinyin-to-character objective is designed to learn useful phonetic representations. In the self-distillation module, the model conducts two forward passes with the phonetics-aware sequence and the raw text as input respectively to obtain two sets of distributions, and the difference between them is minimized by KL-divergence. 

\begin{CJK*}{UTF8}{gbsn}
\paragraph{Phonetics-Aware Input Sequence}
The pinyin of each Chinese character is a sequence of the Latin alphabet and is composed of \textit{initials}, \textit{finals} and \textit{tones} to denote the pronunciation. If characters share the same initial or final, their pronunciations are usually related or similar. In our method, we only consider initials and finals as pinyin information for CSC, as empirically tones are not related to this task.~Given the input $X$, we denote its pinyin sequence as $R=\{(\text{init}_1, \text{final}_1), (\text{init}_2, \text{final}_2),..,(\text{init}_n, \text{final}_n)\}$, where $\text{init}_i$ and $\text{final}_i$ are the initial and final of character $x_i$, respectively. Then, we append $R$ to $X$ and obtain a phonetics-aware sequence $S=\{s_1,s_2,..,s_n, s_{n+1},s_{n+2},..,s_{n+n}\}$ as the final input, where $s_i$ is defined as follows.
\end{CJK*}
\begin{equation}
\small
s_i=\left\{
\begin{array}{lr}
\begin{aligned}
&x_i, & 1\leq &i\leq n \\
\text{init}_{i-n}& , \text{final}_{i-n},& n+1\leq &i\leq n+n 
\end{aligned}
\end{array}
\right..
\end{equation}

\paragraph{Encoder with Separation Mask}
\label{subsec:encoder_with_separation}
We adopt BERT \citep{bert} with a stack of 12 Transformer \citep{transformers} blocks as our encoder. Each Chinese character is encoded as the sum of word embedding, position embedding, and segment embedding. Similarly, the pinyin of each character is encoded as the sum of initial embedding, final embedding, position embedding, and segment embedding, where the position embedding is the same as the character. As a result, the representations of the phonetics-aware input sequence $S$ can be denoted by $H^0=\{h^{0}_{1},h^{0}_{2},..,h^{0}_{n+n}\}$.

The contextual representation of each token is updated by aggregating information from other tokens  via multi-head attention networks ($\text{MHA}$). In the $l$-th layer, the output $O^l$ of each attention head is computed as:
\begin{equation}
\label{eq:mha}
\small
    \begin{aligned}
        Q^l, K^l, V^l &= H^{l-1}W^{l\top}_{Q},  H^{l-1}W^{l\top}_{K},  H^{l-1}W^{l\top}_{V}, \\
        A^{l} &= \text{softmax}( \frac{Q^{l}K^{l\top}}{\sqrt{d}} + M), \\
        O^l &= A^{l}V^{l}. \\
    \end{aligned}
\end{equation}
where $W^{l}_{Q}$, $W^{l}_{K}$, $W^{l}_{V}$ are trainable parameters, $H^{l-1}$ is the output of the previous layer, $d$ is the size of the dimension, and $M$ is a mask matrix.

Specifically, we apply a separation mask to allow for attention from text representations to phonetic representations but not vice versa. Thus, we define the mask matrix $M\in \mathbb{R}^{2n\times 2n}$ in Eq. (\ref{eq:mha}) as: 
\begin{equation}
\small
M_{ij}=\left\{
\begin{array}{lr}
\begin{aligned}
-& \infty, & \text{if } n+1 \leq & i \leq 2n \text{ and } 1 \leq j \leq n \\
&0, &  &\text{otherwise}
\end{aligned}
\end{array}
\right..
\end{equation}
The separation mask ensures that pinyin representations cannot gather information from textual characters when $M_{ij}=-\infty$. Next, $O^l$ from all heads are concatenated then passed through a linear transformation network and a normalization network. After that, the resulting representations are fed into a feed-forward network followed by another normalization network to generate $H^{l}$.  

The final contextual representations $H=\{h_1, h_2,..,h_{n+n}\}$ are produced by taking the last-layer hidden states of the encoder. Then, we compute the probability distribution for the $i$-th character based on $h_i$ by: 
\begin{equation}
\small
    P_i = \text{softmax}(E*h_i + b) \in \mathbb{R}^{|V|}. \label{eq:probability}
\end{equation}
where $E$ is word embedding parameters, $|V|$ denotes the size of vocabulary, and $b$ is a trainable parameter. The prediction loss for the textual part of $S$ is computed as:
\begin{equation}
\small
    \mathcal{L}_{\text{text}}= \frac{1}{n}\sum\limits_{i=1}^{n}-\text{log}P(y_i|S).
\end{equation}

\paragraph{Pinyin-to-Character Objective}
To design the auxiliary pinyin-to-character task, we make a copy of the gold output $Y$ to obtain $Z=\{z_1,..,z_n,z_{n+1},..,z_{n+n}\}$ as the prediction labels of $S$, where $z_1,..,z_n=y_1,..,y_n$ and $z_{n+1},..,z_{n+n}=y_1,..,y_n$. The prediction loss of the pinyin part in $S$ is defined as:
\begin{equation}
\small
    \mathcal{L}_{\text{pinyin}}= \frac{1}{n}\sum\limits_{i=n+1}^{n+n}-\text{log}P(z_{i}|S).
\end{equation}

At inference time, we obtain the prediction $\hat{Y}=\{\hat{y}_1,..\hat{y}_n,\hat{y}_{n+1},..,\hat{y}_{n+n}\}$, where $\hat{y}_i=\text{argmax}(P_i)$. We discard the prediction for the pinyin part and use $\{\hat{y}_1,..\hat{y}_n\}$ as the final output.

\paragraph{Self-Distillation Module}
After obtaining the output distribution for each character by Equation (\ref{eq:probability}), the model conducts another forward pass with the original sequence $X$ as input, giving rise to another output distribution $Q_i  \in \mathbb{R}^{|V|}$ for each character $x_i$. The two sets of distributions are then forced to be close by applying bidirectional KL-divergence:
\begin{equation}
\small
    \mathcal{L}_{kl} = \frac{1}{n}\sum\limits_{i=1}^{n}\frac{1}{2} (\mathcal{D}_{kl}( P_i||Q_i) + \mathcal{D}_{kl}( Q_i|| P_i)).
\end{equation}

Besides, the prediction objective of the second pass is also included in the training:
\begin{equation}
\small
    \mathcal{L}_{\text{raw-text}}= \frac{1}{n}\sum\limits_{i=1}^{n}-\text{log}P(y_i|X).
\end{equation}

\paragraph{Joint Learning}
\label{subsec:joint_learning} To train the model, we combine the phonetics-aware loss and the self-distillation loss into a joint training framework as:
\begin{equation}
\small
    \mathcal{L}= \underbrace{ \mathcal{L}_{\text{text}} + \alpha\mathcal{L}_{\text{pinyin}} }_{\text{phonetics-aware loss}} + \underbrace{ \beta\mathcal{L}_{kl} + \gamma\mathcal{L}_{\text{raw-text}}}_{\text{self-distillation loss}}.
\end{equation}
where $\alpha$, $\beta$, and $\gamma$ are tunable hyperparameters. 

\subsection{Pre-training}
\label{subsec:pre-train}
Pinyin sequences can be regarded as a special form of natural language sequences. Since they are not presented in the original pre-training process of language models, reasonably, they can be pre-trained on large-scale corpora to obtain better pinyin representations for fine-tuning. Therefore, we pre-train DORM on two large corpora, namely wiki2019zh\footnote{\url{https://github.com/brightmart/nlp_chinese_corpus}} and weixin-public-corpus\footnote{\url{https://github.com/nonamestreet/weixin_public_corpus}}. The format of input sequences and the model structure are the same as in fine-tuning. DORM is trained by recovering 15\% randomly selected characters in the input, which were replaced by phonologically similar or random characters. Moreover, the pinyin-to-character objective is also included. More implementation details are given in Appendix~\ref{sec:pretrain_appendix}.

\section{Experiments}
In this section, we introduce the details of our experiments to evaluate the proposed model. 
\subsection{Datasets and Metrics}
We conduct main experiments on three CSC benchmarks, including SIGHAN13 \citep{sighan13}, SIGHAN14 \citep{sighan14}, and SIGHAN15 \citep{sighan15}. Following previous work \citep{pointer-networks, spellgcn, realise}, we merge the three SIGHAN training sets and another 271K pseudo samples generated by ASR or OCR \citep{wang271k} as our training set. We evaluate our model on the test sets of SIGHAN13, SIGHAN14, and SIGHAN15, respectively. Since the original SIGHAN datasets are in Traditional Chinese, they are converted to Simplified Chinese by OpenCC\footnote{\url{https://github.com/BYVoid/OpenCC}}. We adopt the pypinyin toolkit\footnote{\url{https://pypi.org/project/pypinyin/}} to obtain the pinyin of each character.

We use the metrics of sentence-level precision, recall, and F1 to evaluate our model for detection and correction. For detection, all misspelled characters in a sentence should be detected correctly to count it as correct. For correction, a sentence is considered as correct if and only if the model detects and corrects all erroneous characters in this sentence. More details about the datasets and the metrics are presented in Appendix~\ref{sec:dataset_and_metrics_appendix}.

\begin{table*}[h]
\small
    \centering
    \begin{tabular}{c | l | p{1cm}<{\centering} p{1.0cm}<{\centering} p{0.9cm}<{\centering} | p{1cm}<{\centering} p{1cm}<{\centering} p{0.9cm}<{\centering} }
    \toprule
    \multirow{2}{*}{Dataset} & \multicolumn{1}{c|}{\multirow{2}{*}{Methods}} & \multicolumn{3}{c|}{Detection (\%)} & \multicolumn{3}{c}{Correction (\%)} \\
    \cline{3-5} \cline{6-8}
     &  & precision & recall  & F1 & precision & recall & F1 \\
     \hline

     \multirow{7}{*}{SIGHAN15} & BERT      & 74.2 & 78.0 & 76.1 & 71.6 & 75.3 & 73.4 \\
                               & SpellGCN \citep{spellgcn} & 74.8 & 80.7 & 77.7 & 72.1 & 77.7 & 75.9 \\
                               & DCN \citep{dcn}           & 77.1 & 80.9 & 79.0 & 74.5 & 78.2 & 76.3 \\
                      & PLOME \citep{plome}                & 77.4 & 81.5 & 79.4 & 75.3 & 79.3 & 77.2 \\
                      & MLM-phonetics \citep{mlmphonetics} & 77.5 & \underline{83.1} & 80.2 & 74.9 & 80.2 & 77.5 \\
                               & REALISE \citep{realise}   & 77.3 & 81.3 & 79.3 & 75.9 & 79.9 & 77.8 \\
                           & LEAD \citep{li2022dictionary} & \textbf{79.2} & 82.8 & \underline{80.9} & \textbf{77.6} & \underline{81.2} & \underline{79.3} \\
     \cline{2-8}
                               & DORM (ours)                 & \underline{77.9} & \textbf{84.3} & \textbf{81.0} & \underline{76.6} & \textbf{82.8} & \textbf{79.6}   \\
     \hline 
     \hline
     \multirow{7}{*}{SIGHAN14} & BERT      & 64.5 & 68.6 & 66.5 & 62.4 & 66.3 & 64.3 \\
                               & SpellGCN \citep{spellgcn} & 65.1 & 69.5 & 67.2 & 63.1 & 67.2 & 65.3 \\
                               & DCN \citep{dcn}           & 67.4 & 70.4 & 68.9 & 65.8 & 68.7 & 67.2 \\
                      & MLM-phonetics \citep{mlmphonetics} & 66.2 & \textbf{73.8} & 69.8 & 64.2 & \textbf{73.8} & 68.7 \\
                               & REALISE \citep{realise}   & 67.8 & 71.5 & 69.6 & 66.3 & 70.0 & 68.1 \\
                           & LEAD \citep{li2022dictionary} & \textbf{70.7} & 71.0 & \underline{70.8} & \textbf{69.3} & 69.6 & \underline{69.5} \\
     \cline{2-8}
                               & DORM (ours)                 & \underline{69.5} & \underline{73.1} & \textbf{71.2} & \underline{68.4} & \underline{71.9} & \textbf{70.1}   \\
     \hline
     \hline
     \multirow{7}{*}{SIGHAN13} & BERT      & 85.0 & 77.0 & 80.8 & 83.0 & 75.2 & 78.9 \\
                               & SpellGCN \citep{spellgcn} & 80.1 & 74.4 & 77.2 & 78.3 & 72.7 & 75.4 \\
                               & DCN \citep{dcn}           & 86.8 & 79.6 & 83.0 & 84.7 & 77.7 & 81.0 \\
                      & MLM-phonetics \citep{mlmphonetics} & 82.0 & 78.3 & 80.1 & 79.5 & 77.0 & 78.2 \\
                               & REALISE \citep{realise}   & \textbf{88.6} & 82.5 & 85.4 & \underline{87.2} & 81.2 & 84.1 \\
                           & LEAD \citep{li2022dictionary} & \underline{88.3} & \underline{83.4} & \underline{85.8} & \textbf{87.2} & \underline{82.4} & \underline{84.7} \\
     \cline{2-8}
                               & DORM (ours)                 & 87.9 & \textbf{83.7} & \textbf{85.8}   & 86.8   & \textbf{82.7}   & \textbf{84.7}   \\

    \bottomrule
    
    \end{tabular}
    \caption{Overall results of DORM and baselines on SIGHAN13/14/15 in detection/correction precision, recall, and F1. The best results are shown in bold and the second-best results are underlined. The results of baselines are cited from the corresponding papers.}
    \label{tab:overall_results}
    \vspace{-4mm}
    
\end{table*}

\subsection{Baselines}

We compare our DORM with the following baselines. \textbf{BERT} \citep{bert} is initialized with pre-trained $\text{BERT}_{\text{base}}$ and fine-tuned on the training set directly.
\textbf{SpellGCN} \citep{spellgcn} models prior knowledge between phonetically or graphically similar characters with graph convolutional networks.
\textbf{DCN} \citep{dcn} uses a Pinyin Enhanced Candidate Generator to introduce phonological information and then models the connections between adjacent characters.
\textbf{MLM-phonetics} \citep{mlmphonetics} integrates phonetic features during pre-training with a special masking strategy that replaces words with phonetically similar words. \textbf{PLOME} \citep{plome} utilizes GRU networks to model phonological and visual knowledge during pre-training with a confusion set-based masking strategy.
\textbf{REALISE} \citep{realise} learns semantic, phonetic, and visual representations with three encoders and fuses them with a gate mechanism.
\textbf{LEAD} \citep{li2022dictionary} models phonetic, visual, and semantic information by a contrastive learning framework.
Additionally, the implementation details of our DORM are presented in Appendix~\ref{sec:implementation_details_appendix}.

\subsection{Overall Results}
As the overall results show in Table~\ref{tab:overall_results}, the proposed DORM outperforms existing state-of-the-art methods in both detection and correction F1 scores on SIGHAN13/14/15 test datasets, which demonstrates the effectiveness of this model. Compared with other models utilizing phonetic and visual features (e.g., REALISE and PLOME) and models pre-trained on larger corpora (e.g., PLOME and MLM-phonetics), which have access to further external information, DORM still achieves favourable improvement in detection/correction F1. We also note that the improvements in detection/correction recall are prominent and consistent across different test sets. These results suggest that our model is able to capture phonetic information more effectively. Although the improvement in precision is not as encouraging as recall and F1, its performance is still competitive compared with other methods also including phonetic information in this task.

\section{Analysis and Discussion}

In this section, we further analyze and discuss our model quantitatively and qualitatively.

\subsection{Ablation Study}

\begin{table}[t]
\small
\centering
    \resizebox{\columnwidth}{!}{
        \begin{tabular}{l|c c c}
            \toprule
            \multirow{2}{*}{{Method}}  &  \multicolumn{3}{c}{{Correction F1}\ {$(\Delta)$}} \\
            \cline{2-4}
            \\[-0.9em]
                                         & {SIGHAN13}     & {SIGHAN14}        &      {SIGHAN15}   \\
            \hline
            \\[-0.9em]
            \textbf{DORM}               & \textbf{84.7}    &   \textbf{70.1}   &   \textbf{79.6}  \\
            \ \textit{\, w/o} SM                          & 83.6 (-1.1)  & 67.4 (-2.7)  & 79.0 (-0.6) \\
            \ \textit{\, w/o} SD                          & 83.1 (-1.6)  & 69.1 (-1.0)  & 78.9 (-0.7) \\
            \ \textit{\, w/o} $\mathcal{L}_{\text{pinyin}}$  & 84.2 (-0.5) & 68.3 (-1.8)  & 79.2 (-0.4)  \\
            \ \textit{\, w/o} pre-training                & 83.7 (-1.0) & 66.9 (-3.2)  & 78.6 (-1.0) \\
            \hline
            \ \textit{\, w/o} SD\&SM                     & 82.1 (-2.6) & 68.3 (-1.8)  & 77.1 (-2.5) \\
            \ \textit{\, w/o} SD\&$\mathcal{L}_{\text{pinyin}}$           & 83.0 (-1.7)  & 68.7 (-1.4) & 77.8 (-1.8)  \\
            \ \textit{\, w/o} SD\&$\mathcal{L}_{\text{pinyin}}$\&SM       & 81.4 (-3.3)  & 67.3 (-2.8)  & 76.9 (-2.7) \\
           
            \bottomrule
        
        \end{tabular}
    }
    \caption{Results of ablation study in correction F1 on SIGHAN13/14/15, where ``\textit{w/o}'' means without, ``$\mathcal{L}_{\text{pinyin}}$'' means the pinyin-to-character objective, ``SM'' denotes the separation mask, ``SD'' denotes the self-distillation module, and ``$\Delta$'' denotes the change of performance.}
    \label{tab:ablation_study_table}
    \vspace{-2mm}
\end{table}

To investigate the contribution of key components of our model, we ablate them in turn and report the F1 performance for the correction task on SIGHAN13/14/15 in Table~\ref{tab:ablation_study_table}. As shown in the first group, eliminating the separation mask leads to considerable performance declines, showing that preventing pinyin representations from attending to textual information is necessary to learn useful phonetic representations. Moreover, removing self-distillation also leads to performance degradation, which suggests that the module is useful to avoid overfitting pinyin. When $\mathcal{L}_{\text{pinyin}}$ is discarded, the performance drops correspondingly, meaning that phonetic features tend to be ignored without the pinyin-to-character objective. Moreover, a sharp decline is observed when dropping the pre-training phase, which implies that pre-training on large-scale corpora indeed improves phonetic representations. More experimental results of various combinations in the second group further reveal the contribution of these components.

\subsection{Effect of Phonetic Knowledge}
According to the assumption, more phonetically similar misspellings should be restored with the assistance of phonetic knowledge. To show this, we focus on the recall performance of different models on phonetically misspelled characters of SIGHAN13/14/15. We collect 1130/733/668 such misspellings from the three test sets, accounting for about 93\%/95\%/95\% of all misspellings, respectively. From the results in Table~\ref{tab:epk}, we can note that our model achieves 93.5\%/82.1\%/90.0\% recall scores and outperforms two phonetic-based models (i.e., SCOPE \citep{scope} and REALISE) consistently. In particular, it beats BERT by a large margin. These results indicate that phonetic knowledge is essential to CSC and our model is able to utilize phonetic knowledge more effectively.

\begin{table}[h]
\centering
    \resizebox{\columnwidth}{!}{
        \begin{tabular}{l|c c c}
            \toprule
            \multirow{2}{*}{Model}  &  \multicolumn{3}{c}{Recall (\%)} \\
            \cline{2-4}
                         & SIGHAN13  & SIGHAN14  & SIGHAN15   \\
            \hline
            DORM     &  \textbf{93.5}   &  \textbf{82.1} &  \textbf{90.0}   \\
            ${\text{SCOPE}}^{\dagger}$      &  91.6  &  80.2  &  87.6    \\
            ${\text{REALISE}}^{\dagger}$    &  89.8  &  78.2  &  84.7    \\
            BERT &  88.8  & 75.2  & 82.8 \\
            \bottomrule
        
        \end{tabular}
    }
    \caption{The performance of models in restoring phonetically misspelled characters on SIGHAN13/14/15. Results marked with ``$\dagger$'' are obtained by executing released models from corresponding papers.}
    \label{tab:epk}
    \vspace{-2mm}
\end{table}

\subsection{Effect of Self-Distillation}
The self-distillation module is introduced for DORM to avoid overfitting pinyin information. To show the effect of this module, we record the number of normal characters that are mistakenly treated as misspellings (i.e., overcorrections), as well as the number of misspellings not restored (i.e., undercorrections) in the three test sets. The results in Table~\ref{tab:esd} show that the number of undercorrections is significantly reduced when phonological information but not self-distillation is introduced, while the number of overcorrections generally stays unchanged except on SIGHAN13. These results demonstrate that after including the self-distillation module, the numbers of overcorrections and undercorrections are both reduced compared with the baseline, demonstrating that self-distillation indeed alleviates the overfitting issue.

\begin{table}[h]
\centering
    \resizebox{\columnwidth}{!}{
        \begin{tabular}{l|c c c}
            \toprule
            \multirow{2}{*}{Model}  &  \multicolumn{3}{c}{ {\#Overcorrections}/{\#Undercorrections}} \\
            \cline{2-4}
                         & SIGHAN13  & SIGHAN14  & SIGHAN15   \\
            \hline
            $\text{BERT}$        &  {103}/129  & 175/177  & 120/106 \\
            DORM \textit{w/o} SD  &  118/{75}  & 172/{134}  & 119/{63} \\
            DORM                  &  107/77  & {161}/136  & {116}/65 \\
            \bottomrule
        
        \end{tabular}
    }
    \caption{The effect of self-distillation in reducing overcorrections and undercorrections on SIGHAN13/14/15. ``\textit{w/o} SD'' means without the self-distillation module.}
    \label{tab:esd}
    \vspace{-2mm}
\end{table}

\begin{CJK*}{UTF8}{gbsn}
\begin{figure}[htb]
    \centering
    \includegraphics[width=\columnwidth]{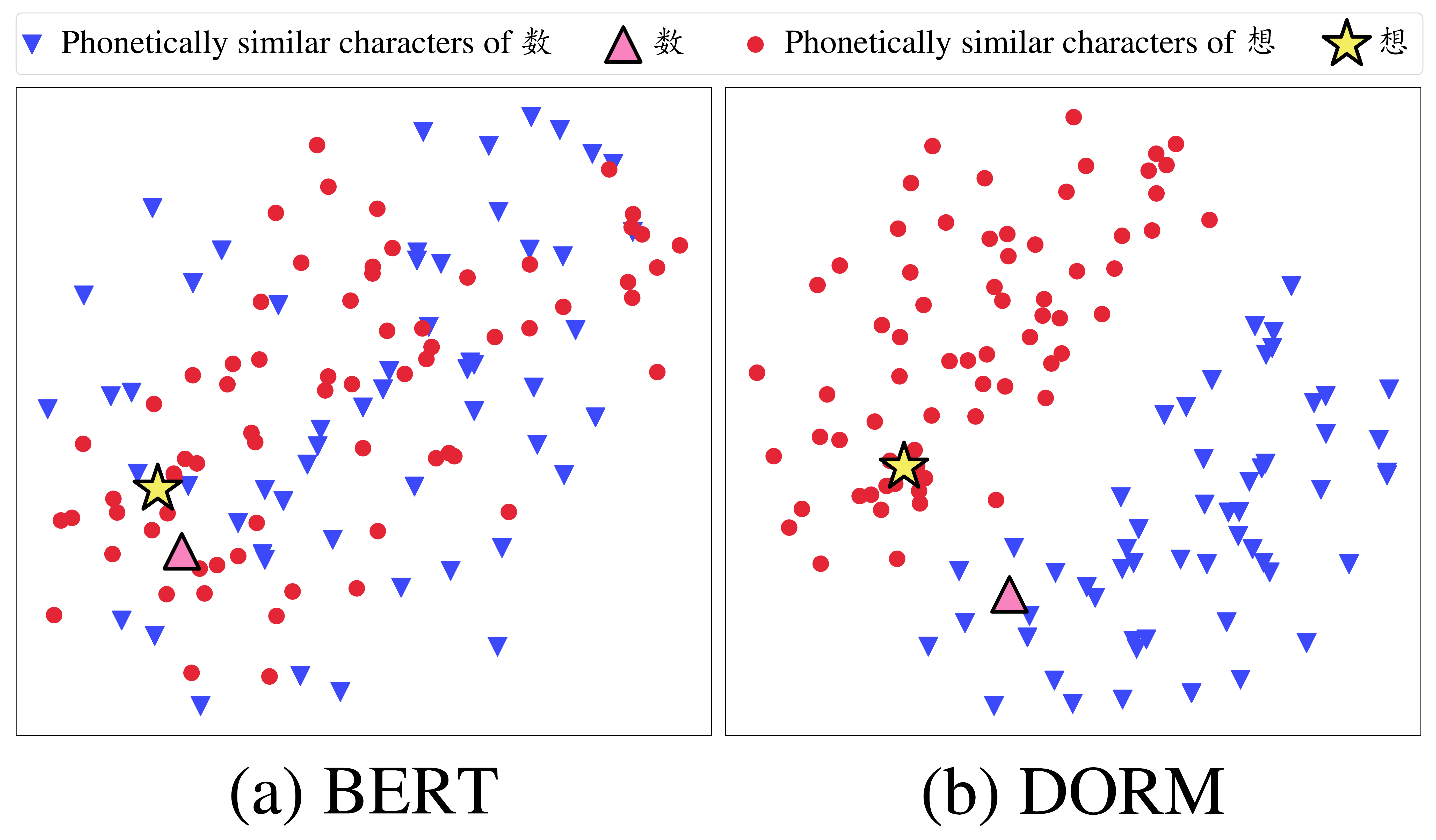}
    \caption{Visualization of character representations, in which (a) is fine-tuned BERT and (b) is our DORM. Two pivot characters ``数'' (number) and ``想'' (want) have different pronunciations.}
    \label{fig:tsne_visualization}
    \vspace{-3mm}
\end{figure}
\end{CJK*}

\begin{figure*}[htb]
    \centering
    \includegraphics[width=0.95\textwidth]{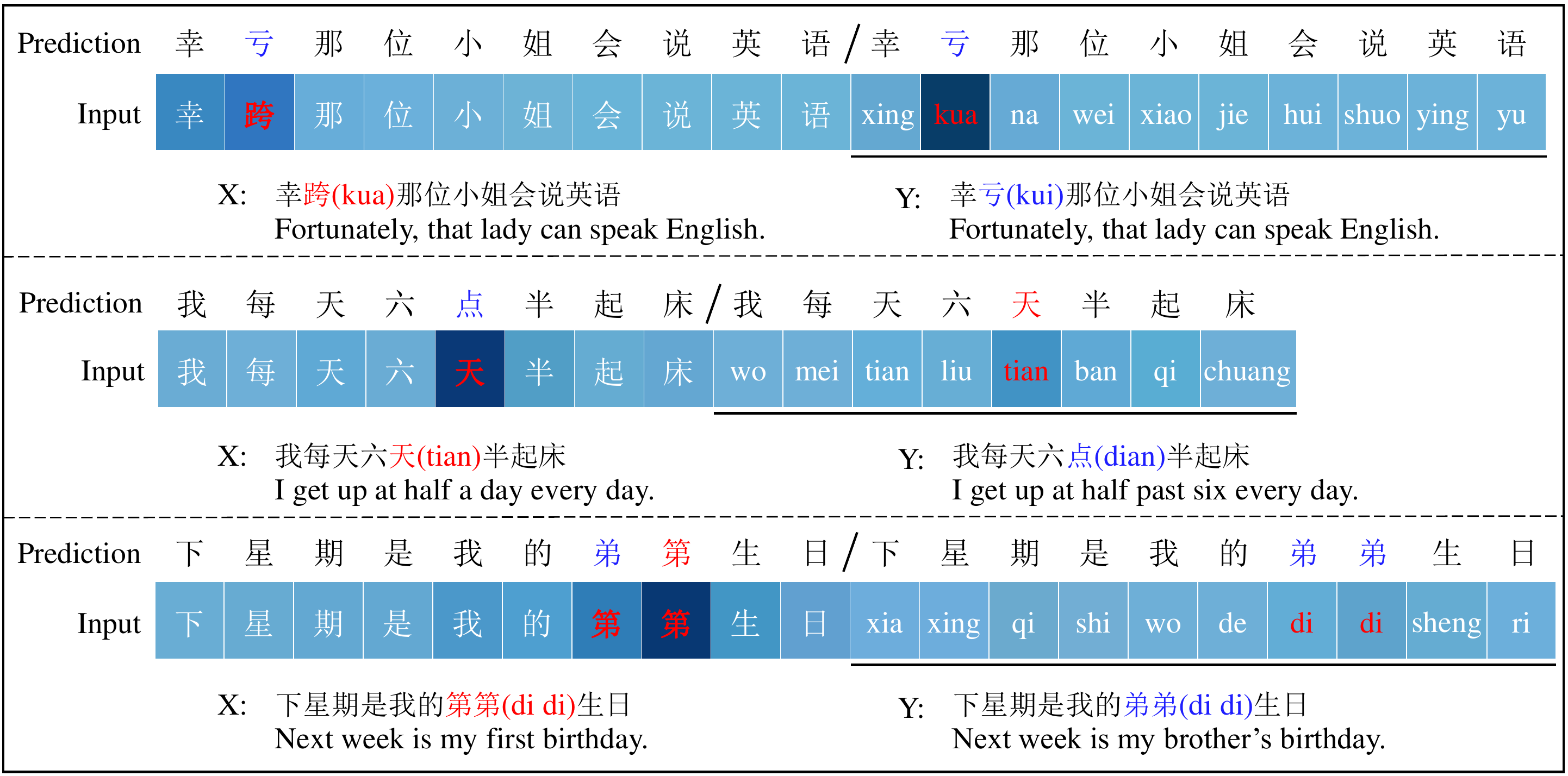}
    \caption{Case study on SIGHAN15, where misspellings and corresponding answers are highlighted in red and blue, respectively. The phonetic input is underlined and its prediction is discarded during inference. Attention weights from misspellings to the input sequence are also visualized where darker colors mean larger weights.}
    \label{fig:attention_case_study}
    \vspace{-3mm}
\end{figure*}

\subsection{Visualization}
\begin{CJK*}{UTF8}{gbsn}
Ideally, the introduction of phonetic knowledge should improve Chinese character representations in that phonetically similar characters are pulled closer in the space. To show the effect, we employ t-SNE \citep{tsne} to visualize character representations generated by our model, with fine-tuned BERT as the baseline. We randomly select two characters ``数'' and ``想'' of different pronunciations and collect about 60 phonetically similar characters provided by \citet{sighan13} for eacknow.
We plot the two groups of representations in Figure \ref{fig:tsne_visualization}, from which we can note that the representations produced by fine-tuned BERT are scattered and less distinguishable between the groups. However, our model separates them into two distinct clusters according to the pivot characters, demonstrating that our model can better model the relationships among phonetically similar characters for CSC. 
\end{CJK*}

\subsection{Case Study}

Finally, we provide a case study with two good and one bad examples to analyze our model. We visualize the attention weights from each misspelled character to the other positions in the phonetics-aware sequence to show how our model utilizes phonetic information. As presented in Figure \ref{fig:attention_case_study}, in the first case both the textual and phonetic parts make correct predictions. After looking into the attention weights, we note the prediction for the misspelled position pays much attention to its previous position, the current position, and its pinyin position.  In the second case, while the phonetic part leads to a wrong prediction, our model focuses more on the textual part and eventually makes a correct prediction.  In the third case, although the prediction of the pinyin part is accurate, the textual part fails to pay much attention to it and causes a wrong prediction, suggesting that there is still room for improvement in balancing phonetic and semantic information. These cases intuitively show how our model uses phonetic information to correct misspelled characters. 

\vspace{-1mm}
\section{Conclusion}
\vspace{-1mm}

In this paper, we propose DORM in an attempt to improve the effect of using phonetic knowledge in Chinese Spelling Correction (CSC). To this end, we propose to disentangle textual and phonetic features and construct a phonetics-aware input to allow for direct interaction between them. We also introduce a pinyin-to-character objective to force the model to recover the correct characters based solely on pinyin information, where a separation mask is applied to prevent exposing textual information to phonetic representations. Besides, we propose a novel self-distillation module for DORM to avoid overfitting pinyin information. Extensive experiments on three widely-used CSC datasets show that this model outperforms existing state-of-the-art baselines. Detailed analysis and studies show that direct interaction between characters and pinyin is beneficial to better restore misspelled characters. Through this work, we demonstrate the merit of disentangling phonetic features from textual representations when solving CSC.

\section*{Acknowledgements}
We appreciate the anonymous reviewers for their valuable comments.~This work was supported by the National Natural Science Foundation of China (No.~62176270), the Guangdong Basic and Applied Basic Research Foundation (No. 2023A1515012832), and the Program for Guangdong Introducing Innovative and Entrepreneurial Teams (No. 2017ZT07X355).

\section*{Limitations}

The potential limitations of our model are threefold. First, the training process requires more computational cost as the model needs to conduct two forward passes for each sample in the self-distillation module. Second, there is still room for improvement to reduce the model's overcorrection of legal characters. Third, the phonetics-aware sequence doubles the length of the original input, which demands extra computation cost at inference time.



\section*{Ethics Statement}
This work aims to propose a technical method to utilize phonetic knowledge more effectively for Chinese Spelling Correction, which does not involve ethical issues. The datasets used in this work are all publicly available.

\bibliography{anthology,custom}
\bibliographystyle{acl_natbib}

\appendix

\section{Pre-training}
\label{sec:pretrain_appendix}

There are 1 million and 0.7 million articles in wiki2019zh corpus and weixin-public-corpus, respectively. First, we generate continuous sentence fragments of at most 256 characters from two corpora as pre-training samples. Then, we randomly sample 15\% characters in each fragment and replace them with: (1) a phonologically similar character 80\% of the time, (2) a randomly selected character 10\% of the time, and (3) unchanged 10\% of the time. After that, we acquire the pinyin sequence of the corrupted fragment and construct a phonetics-aware sequence, and replicate the original fragment to construct the prediction labels. We obtain a total of 4.8 million samples for pre-training. 

The architecture of the model for pre-training is the same as described in Section \ref{sec:model_detail}. The model is trained by recovering those selected characters from the phonetics-aware sequence and the pinyin-to-character objective, while the self-distillation module is not required. The batch size is set to 72 and the learning rate is 5e-5.

\section{Datasets and Evaluation Metrics}
\label{sec:dataset_and_metrics_appendix}
\begin{CJK*}{UTF8}{gbsn}
The statistics of the training and test datasets for the experiments are presented in Table~\ref{tab:statistics_of_datasets}. It is worth mentioning that we post-process the predictions of characters ``的'', ``得'' and ``地'' on the SIGHAN13 test set following previous work \citep{realise}, because the annotations for these characters are not accurate. Specifically, the detection and correction of the three characters are not considered.

\begin{table}[h]
\centering
    \resizebox{\columnwidth}{!}{
        \begin{tabular}{l r r r}
            \toprule
            Train   & \#Sent  &  \#Errors     &  Avg. Length   \\
            \hline
            SIGHAN15        & 2,338    & 3,037   & 31.3 \\
            SIGHAN14        & 3,437    & 5,122   & 49.6 \\
            SIGHAN13        & 700      & 343     & 41.8 \\
            271K pseudo data & 271,329  & 381,962  & 42.6 \\
            \hline
            \hline
            Test   & \#Sent  &  \#Errors     &  Avg. Length   \\
            \hline
            SIGHAN15        & 1,100    & 703       & 30.6 \\
            SIGHAN14        & 1,062    & 771       & 50.0 \\
            SIGHAN13        & 1,000    & 1,224     & 74.3 \\
            \bottomrule
        
        \end{tabular}
    }
    \caption{Statistics of the SIGHAN training and test datasets. We train our model on the combination of all the training sets and evaluate it on each test dataset.}
    \label{tab:statistics_of_datasets}
\end{table}
\end{CJK*}

\section{Implementation of DROM}
\label{sec:implementation_details_appendix}
Our encoder contains 12 attention heads with a hidden size of 768 (about 110M parameters) and is initialized with weights from Chinese BERT-wwm \citep{bert_wwm}. The embeddings of initials and finals are randomly initialized. Our model is firstly pre-trained and then fine-tuned on the CSC training set. We apply the AdamW optimizer \citep{adamw} to fine-tune the model for 3 epochs on three 24G GeForce RTX 3090 GPUs. The learning rate is scheduled to decrease gradually after linearly increasing to 75e-6 during warmup. The maximum sentence length is set to 140. The batch sizes for training and evaluation are set to 48 and 32, respectively. The hyperparameters of $\alpha$, $\beta$, and $\gamma$ are set to 1, 1.2 and 0.97, respectively. Our implementation is based on Huggingface's Transformer \citep{huggingface} in PyTorch.

\end{document}